\documentclass{article}

\usepackage[preprint]{neurips_2026}

\usepackage[T1]{fontenc}    
\usepackage{hyperref}       
\usepackage{url}            
\usepackage{booktabs}       
\usepackage{amsmath}        
\usepackage{amsfonts}       
\usepackage{amssymb}        
\usepackage{nicefrac}       
\usepackage{microtype}      
\usepackage[table]{xcolor}  
\usepackage{caption}        
\usepackage{wrapfig}        

\usepackage{multirow}
\usepackage{algorithm,algpseudocode}
\usepackage{colortbl}
\usepackage{pgf}
\usepackage{graphicx}

\IfFileExists{twemojis.sty}{\usepackage{twemojis}}{}
\colorlet{darkgreen}{green!65!black}

\makeatletter
\@ifpackageloaded{twemojis}{%
  \newcommand{\heartemoji}{$^{\twemoji{red heart}}$}%
}{%
  \newcommand{\heartemoji}{}%
}
\makeatother

\title{FacePhys: State of the Heart Learning\heartemoji}

\author{%
  \textbf{Jiankai Tang}$^{1,2,*}$ \quad
  \textbf{Kegang Wang}$^{1,2,*}$ \quad
  \textbf{Yuntao Wang}$^{1,3,4,\dagger}$ \quad
  \textbf{Xin Liu}$^{5}$ \\
  \textbf{Yuxuan Fan}$^{1}$ \quad
  \textbf{Jiatong Ji}$^{1}$ \quad
  \textbf{Yuanchun Shi}$^{1,3,4}$ \quad
  \textbf{Daniel McDuff}$^{5,\dagger}$ \\[1ex]
  $^{1}$Tsinghua University \quad
  $^{2}$FacePhys \\
  $^{3}$National Key Laboratory of Human Factors Engineering \\
  $^{4}$Qinghai University \quad
  $^{5}$University of Washington
}

\begin{document}

\maketitle

\begingroup
\renewcommand{\thefootnote}{}
\footnotetext{%
  $^{*}$Equal contribution
  \quad
  $^{\dagger}$Corresponding authors (yuntaowang@tsinghua.edu.cn)
}
\endgroup
\begin{abstract}
Vital sign measurement using cameras presents opportunities for comfortable, ubiquitous health monitoring. Remote photoplethysmography (rPPG), a foundational technology, enables cardiac measurement through minute changes in light reflected from the skin. However, practical deployment is limited by the computational constraints of performing analysis on front-end devices and the accuracy degradation of transmitting data through compressive channels that reduce signal quality. We propose a memory efficient rPPG algorithm - \emph{FacePhys} - built on a Linear Time-Invariant (LTI) state space model with temporal-spatial duality, which resolves the trilemma of model scalability, cross-dataset generalization, and real-time operation. Leveraging a transferable heart state, FacePhys captures subtle periodic variations across video frames through streaming per-frame waveform inference, enabling training on extended video sequences and supporting low-latency operation.
FacePhys establishes a new state-of-the-art, with a substantial 49\% reduction in heart rate estimation error. Our solution enables streaming waveform inference with a memory footprint of 3.6 MB and per-frame latency of 9.46 ms -- surpassing existing methods by 83\% to 99\%. The model and real-time demo are available at \href{https://github.com/FacePhys/FacePhys-Release}{GitHub} and \href{https://facephys.com}{facephys.com}.


    
\end{abstract}

\vspace{-0.5em}
\section{Introduction}
\label{sec:intro}
\vspace{-0.3em}

Remote photoplethysmography (rPPG) is a non-contact technique for measuring cardiac activity by analyzing light reflected from skin~\citep{mcduff2023camera}. The subtle pulsatile variations are typically not visible to the naked eye, but they can be captured by cameras and analyzed to recover heart rate~\citep{wu2012eulerian}, heart rate variability and other vital signs that can be derived from the waveform (e.g., blood pressure~\citep{curran2023camera}). The attraction of non-invasive measurement of cardiac activity has led to rPPG technology being integrated into a number of applications, including health monitors~\citep{lu2021dual,liu2024summit}, emotion recognition systems~\citep{wang2023physbench,song_pulsegan_2020}, and fatigue detection models~\citep{du2022multimodal}.

Deep learning is the predominant and superior approach used in rPPG models~\citep{mcduff2023camera}, which can be effective at disentangling the cardiac pulse from head motions, ambient lighting changes, and other confounders~\citep{tang2023mmpd,nowara_benefit_2020}. However, there are several reasons that performing inference \emph{on-device} is very desirable, and this means high computational efficiency \emph{and} accuracy are both important.

\textbf{First}, the PPG is degraded more than other information by video compression~\citep{mcduff2017the} meaning that it is optimal to process the video as close to the source as possible (i.e., on-device) rather than send it over a bandwidth limited network.
\textbf{Second}, application of rPPG in resource limited settings, such as by mobile health workers in remote settings, means it may not be possible to rely on a cloud service.
\textbf{Third}, cardiac measurement is a highly privacy sensitive application, with videos often containing a patient's face and sensitive physiological signals. Further adding to why streaming and uploading to a server is not ideal.
\textbf{Fourth}, the ability to run at a high frame rates enables opportunistic sensing (e.g., obtaining measurements each time you look at your phone) and helps capture waveform dynamics that could be used to detect atrial fibrillation~\citep{liu2022vidaf} or hypertension~\citep{Yoshioka_2020_CVPR_Workshops} where high-frame rates (at least 100Hz) are a requirement to yield precise measurements.
However, rather than leading to a decrease, the use of deep learning has meant that the computational resources required to run rPPG algorithms has \emph{increased}, making practical applications and real-time inference challenging. 



Some studies attempt to address efficiency issues by reducing the resolution of frames~\citep{rtrppg} or simplifying the model complexity~\citep{efficientphys}; however, these design choices invariably impact accuracy.
Creating smaller, more memory efficient models~\citep{he2019device,lin2019tsm} often requires slicing videos into shorter segments (e.g., 180 frames, approximately 6 seconds)~\citep{liu2022rppgtoolbox,kuang2023shuffle}, an approach that overlooks the importance of longer-term temporal dependencies, leading to the loss of important information during training and inference, and making algorithms more susceptible to noise.

\begin{figure}[t]
    \centering
    \includegraphics[width=1.0\linewidth]{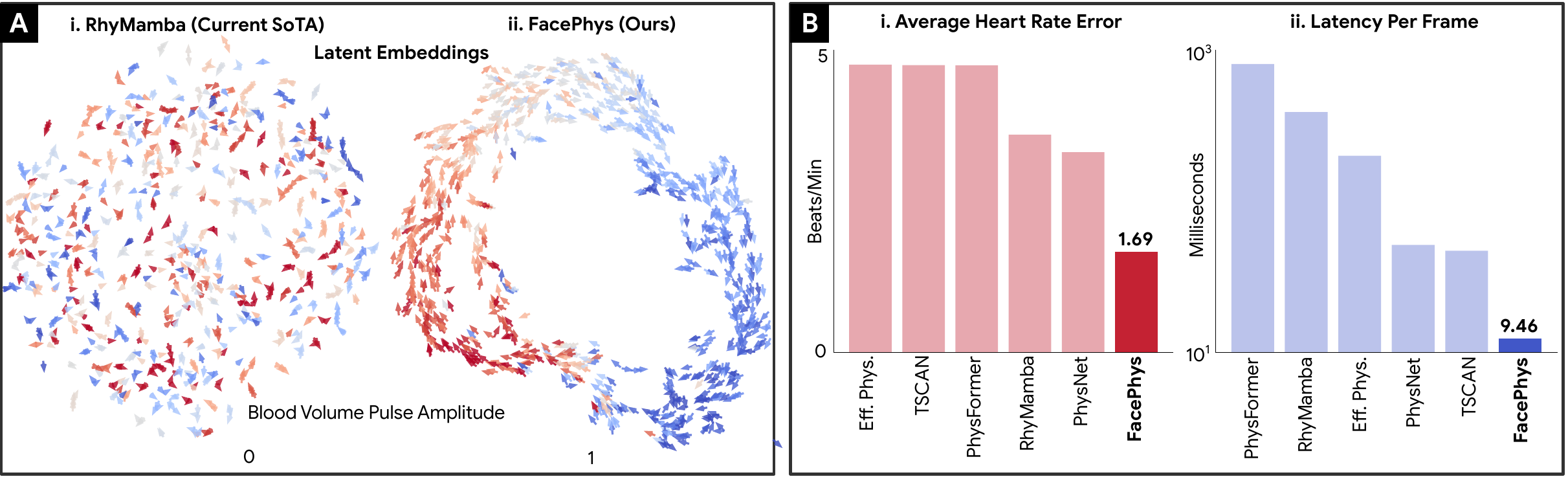}
    \caption{\textbf{The FacePhys State Space Model}, provides an effective representation of the cyclical nature of heart beats, as shown by the latent embeddings (A), combining high accuracy (B-i) and high efficiency (B-ii). We achieve gains of 49\% in heart rate estimation and 83\% in per-frame latency compared to the current state-of-the-art.}
    \vspace{-0.5em}
    \label{fig:teaser}
\end{figure}

This paper proposes an accurate and memory-efficient rPPG model (\textbf{FacePhys}) built on a Linear Time-Invariant (LTI) state space model. FacePhys captures subtle spatial and temporal color changes corresponding to blood volume pulse, enabling accurate prediction of model states and the cardiac signal across arbitrarily long sequences. Leveraging temporal-spatial state space attention duality (TSD), FacePhys supports streaming per-frame waveform inference through recurrent state updates, significantly reducing computational load and achieving real-time operation. Heart rate is computed from the predicted waveform using standard spectral analysis. By learning a strong internal representation, FacePhys demonstrates superior performance across all datasets compared to state-of-the-art (SOTA) methods (see Fig.~\ref{fig:teaser}). The three core contributions of this paper are as follows:
\begin{enumerate}
    \item We built FacePhys, an LTI state space model with TSD that supports streaming \emph{per-frame waveform} inference while preserving long-range periodic information.
    \item  We developed a discrete-time formulation that cuts the model memory to \textbf{3.6\,MB} (up to \textbf{98.4\%} reduction) and achieves sub-\textbf{10\,ms} per-frame latency, verified on mobile browsers.
    \item We evaluated FacePhys on large-scale datasets and demonstrated that it outperforms SOTA methods, including a \textbf{42.3\%} improvement on the challenging MMPD dataset and up to \textbf{49\%} overall improvement.
\end{enumerate}

\vspace{-0.5em}
\vspace{-0.5em}
\section{Related Work}
\label{sec:related}
\vspace{-0.3em}

\label{sec:related:ppg}
\textbf{Efficient rPPG Methods.} Remote physiological sensing has evolved from handcrafted signal processing techniques such as POS~\citep{pos} to deep learning methods~\citep{Gideon2021TheWT,Li2023ContactlessPE}. Early CNN-based rPPG models extract features from facial videos using convolutions and pooling. Seq-rPPG~\citep{wang2023physbench} employs 1D-CNN on temporal features, achieving stable performance with minimal 8x8 facial regions. DeepPhys~\citep{chen2018deepphys} and TS-CAN~\citep{liu2020multi} use 2D-CNNs with temporal shift modules for improved heart rate estimation. PhysNet~\citep{physnet} and iBVPNet~\citep{joshi2024ibvp} leverage 3D-CNNs for spatio-temporal feature extraction. PHASE-Net~\citep{Zhao_2026_CVPR} derives a physics-informed temporal model and uses a causal gated temporal convolutional network for pulse recovery. Despite improvements, CNNs struggle with long-range dependencies, limiting effectiveness in complex scenarios~\citep{tang2025m3pd}. Transformer-based methods address this via self-attention: PhysFormer~\citep{yu2022physformer} focuses on temporal features through long-range dependency modeling, while RhythmFormer~\citep{Zou2024RhythmFormerER} introduces hierarchical periodicity. However, Transformers' quadratic complexity limits practical deployment. Recent Mamba-based approaches~\citep{luo2024physmamba,wu2025cardiacmamba,zou2024rhythmmamba,yan2024physmamba} improve efficiency but remain challenging for embedded devices, motivating more efficient architectures. 
\begin{wraptable}{r}{0.6\textwidth}

     \centering
     \caption{Computational Efficiency on CPUs}
     \label{tab:efficiency}
     \vspace{-0.3cm}
     \begin{tabular}{lcccc}
     \toprule[1.5pt]
     Method & Params.$\downarrow$ & Mem.$\downarrow$  & Latency$\downarrow$ & FLOPs$\downarrow$ \\
     \midrule \midrule
     EFFPhys & 2,189 & 609 & 371 & 39.0 \\
     TSCAN & 533 & 228 & 55.2 & 8.07 \\
     PhysNet & 770 & 230 & 62.1 & 6.16 \\
     PhysFormer & 7,395 & 576 & 2330 & 95.7 \\
     RhyMamba & 4,936 & 347 & 889 & 27.4 \\
     \rowcolor{gray!30}
     FacePhys & 719 &3.6 & 9.46 & 0.15 \\
     \bottomrule[1.5pt]
     \end{tabular}
     \scriptsize \noindent \\Params. (K) = Number of parameters. Mem (MB) = Memory usage. Latency (ms) = Inference time on CPUs. FLOPs (G) = Floating point operations.
     \vspace{-1.5em}
     \end{wraptable}

\textbf{State Space Models (SSMs).} Early SSMs such as HiPPO~\citep{gu2020hipporecurrentmemoryoptimal} and S4~\citep{s4gu2022efficientlymodelinglongsequences} achieve linear-time complexity via continuous-time systems discretized for sequential data, compared to Transformers' quadratic cost. Variants like S5~\citep{s5smith2023simplifiedstatespacelayers}, H3~\citep{fu2023hungryhungryhipposlanguage}, Mamba~\citep{gu2023mamba}, and Mamba-2~\citep{dao2024transformers} enhance capacity and extend to vision. Mamba introduces selective state updates, while Mamba-2 establishes the Structured State Space Duality (SSD) framework unifying SSMs and Transformers~\citep{dao2024transformers}. In vision, VideoMamba achieves strong performance on video understanding~\citep{li2024videomamba}, with Mamba variants succeeding across diverse tasks~\citep{xu2024visual}. We propose FacePhys as an rPPG-specific SSM leveraging SSD for streaming waveform inference with physiologically-motivated periodic attention.

\vspace{-0.5em}
\section{Method}
\label{sec:method}
\vspace{-0.3em}

\subsection{Basic Model}
\begin{figure}[t]
  \vspace{-0.5em}
  \centering
  \begin{minipage}{0.4\linewidth}
    \centering
    \includegraphics[width=\linewidth]{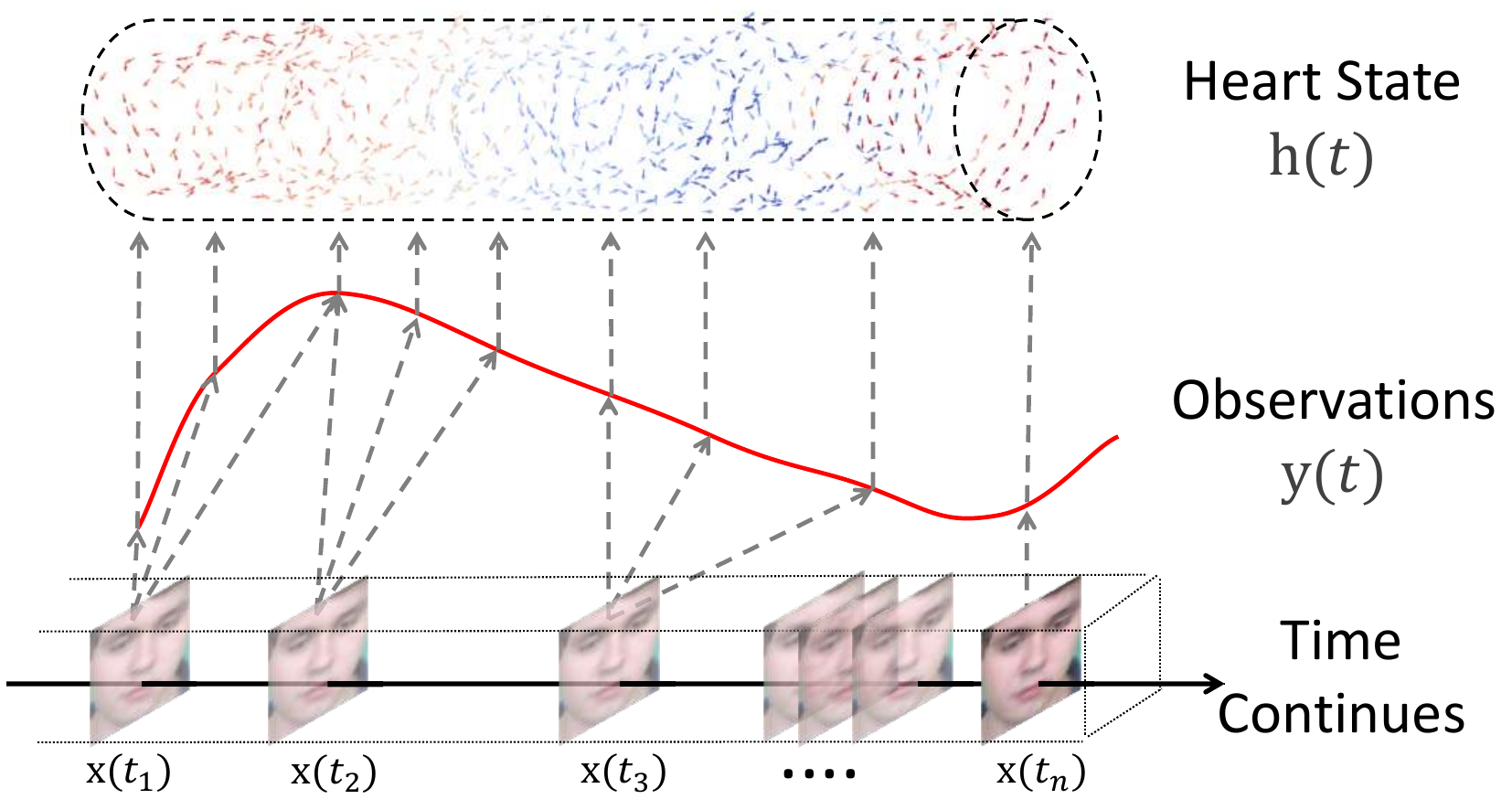}
    \caption{\textbf{LTI heart state.} The time-continuous heart state is discretized as a state space model for efficiency.}
    \label{fig:ode_overview}
  \end{minipage}%
  \hfill
  \begin{minipage}{0.58\linewidth}
    \centering
    \includegraphics[width=\linewidth]{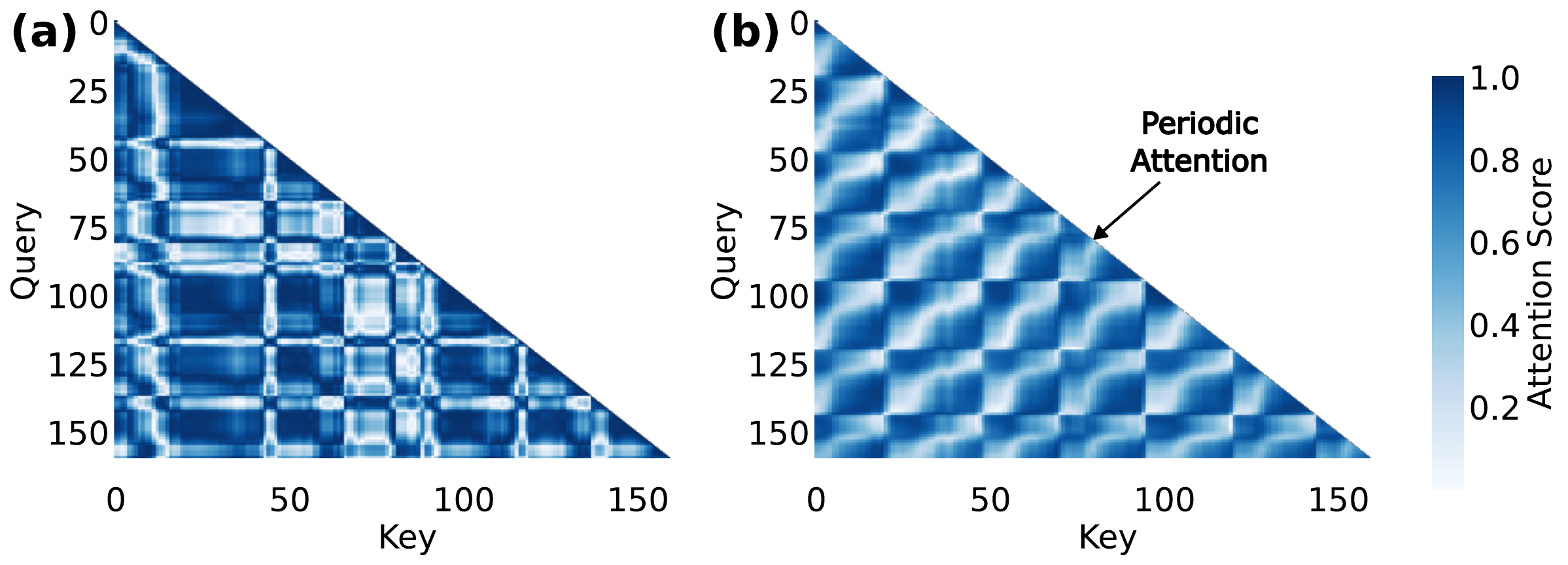}
    \caption{\textbf{Periodic attention.} Introducing trainable complex numbers into the diagonalized state transition matrix generates oscillatory terms for periodic attention.}
    \label{fig:matrix}
  \end{minipage}
\end{figure}

\vspace{-0.2em}
FacePhys extracts the heart state from continuous input $x(t)$, as shown in Fig.~\ref{fig:ode_overview}. The remote PPG signal $y(t)$ can be regarded as an observation of the heart state $h(t)$. The entire model can be structured as a controlled Linear Time-Invariant (LTI) system, as in Equation \eqref{eq:state space}.

\vspace{-0.3em}

\begin{equation}
\begin{aligned}
h'(t) &= \mathbf{A}h(t) + \mathbf{B}x(t) \\
y(t) &= \mathbf{C}h(t) + \mathbf{D}x(t)
\end{aligned}
\label{eq:state space}
\end{equation}
\vspace{-0.2em}

where trainable $\mathbf{A}$ is the state transition matrix, $\mathbf{B}$ is the input matrix, $\mathbf{C}$ is the output matrix, $\mathbf{D}$ is the feedforward matrix, $x(t)$ is the control term (face information), $h(t)$ is the heart state, and $y(t)$ is the observation term (rPPG measurement).

We treat the underlying blood volume pulse as a latent dynamical system rather than a pure temporal dependency, and thus the state transition matrix encodes physiological periodicity instead of generic recurrence. As reported by PhysNet~\citep{physnet}, the RNN variant underperforms its CNN counterpart; we therefore adopt ODEs as a continuous-time state-space formulation better aligned with physiological dynamics than discrete-time RNNs.

\vspace{-0.3em}
\subsection{State Space Duality}
\vspace{-0.2em}

State space duality~\citep{dao2024transformers} is an established method for introducing efficient attention. Solving the continuous-time system in Eq.~\eqref{eq:state space} requires discretization, as real videos consist of discrete frames. Following Mamba-2~\citep{dao2024transformers}, we use zero-order hold (ZOH) discretization to obtain a numerically stable and efficient SSM with strong periodic long-range temporal modeling while retaining the continuous-time inductive bias.

Using the ZOH discretization method, any Linear Time-Invariant (LTI) system can be discretized as follows: $\bar{A} = e^{A \Delta t}$ and $\bar{B} = \left( \int_0^{\Delta t} e^{A \tau} d\tau \right) B$. This property ensures the equivalence of state transitions between the continuous and discrete systems, forming the core theorem of SSD.

\vspace{-0.2em}
The recursive relationship of the hidden state $h_t$ is transformed into matrix multiplication via global convolution expansion: $h_t = \sum_{k=1}^t \bar{A}^{t-k} \bar{B} x_k$. By constraining matrix $\bar{A}$ to be a diagonal matrix, matrix multiplication and inversion can be avoided. The global hidden state $H$, composed of hidden states from all time steps, is expressed as \eqref{eq:global state}. Here, $L$ is a lower triangular matrix, whose elements are defined as $L_{i,j} = \bar{A}^{i-j}$ when $i \geq j$, and $\odot$ denotes the Hadamard product.

\vspace{-0.5em}
\begin{equation}
H = (L \odot (C \bar{B}^{\top})) \cdot X
\label{eq:global state}
\end{equation}
\vspace{-0.5em}

By making matrices $C$ and $\bar{B}$ dependent on the input, this formulation is equivalent to linear attention, expressed as $(L \odot (Q K^T)) \cdot V$. Thus, SSD exhibits long-range modeling capabilities similar to those of transformers.

\begin{wraptable}{r}{0.6\textwidth}
  \vspace{-1.5em}
  \centering
  \caption{\textbf{Memory Efficiency.} Comparison Between TSD and Traditional Structures.}
  \label{tab:memory complexity}
  \begin{tabular}{ccc}
    \toprule[1.5pt]
    Structure       & Training Mem. & Inference Mem. \\
    \midrule \midrule
    3D CNN      &  $O(Tk^3D)$   & $O(k^3D)$ \\
    Transformer      & $O(T^2D)$         & $O(TD)$     \\
    TSD (Ours)      & $O(TD)$         & $O(D+L)$     \\
    \bottomrule[1.5pt]
  \end{tabular}

  \scriptsize \noindent \\ $L$ = SSD state dim. $k$ = 3D kernel size. $D$ = Feature dim. $T$ = Time duration.
  \vspace{-1em}
\end{wraptable}

Due to the step-by-step inference characteristics of SSMs, linear complexity can be achieved during inference as shown in Table~\ref{tab:memory complexity}, which enables efficient solving of the heart state ODE.

\vspace{-0.3em}
\subsection{General Framework}
\vspace{-0.2em}
\begin{figure*}[h]
    \vspace{-0.5em}
    \centering
    \includegraphics[width=1.0\textwidth]{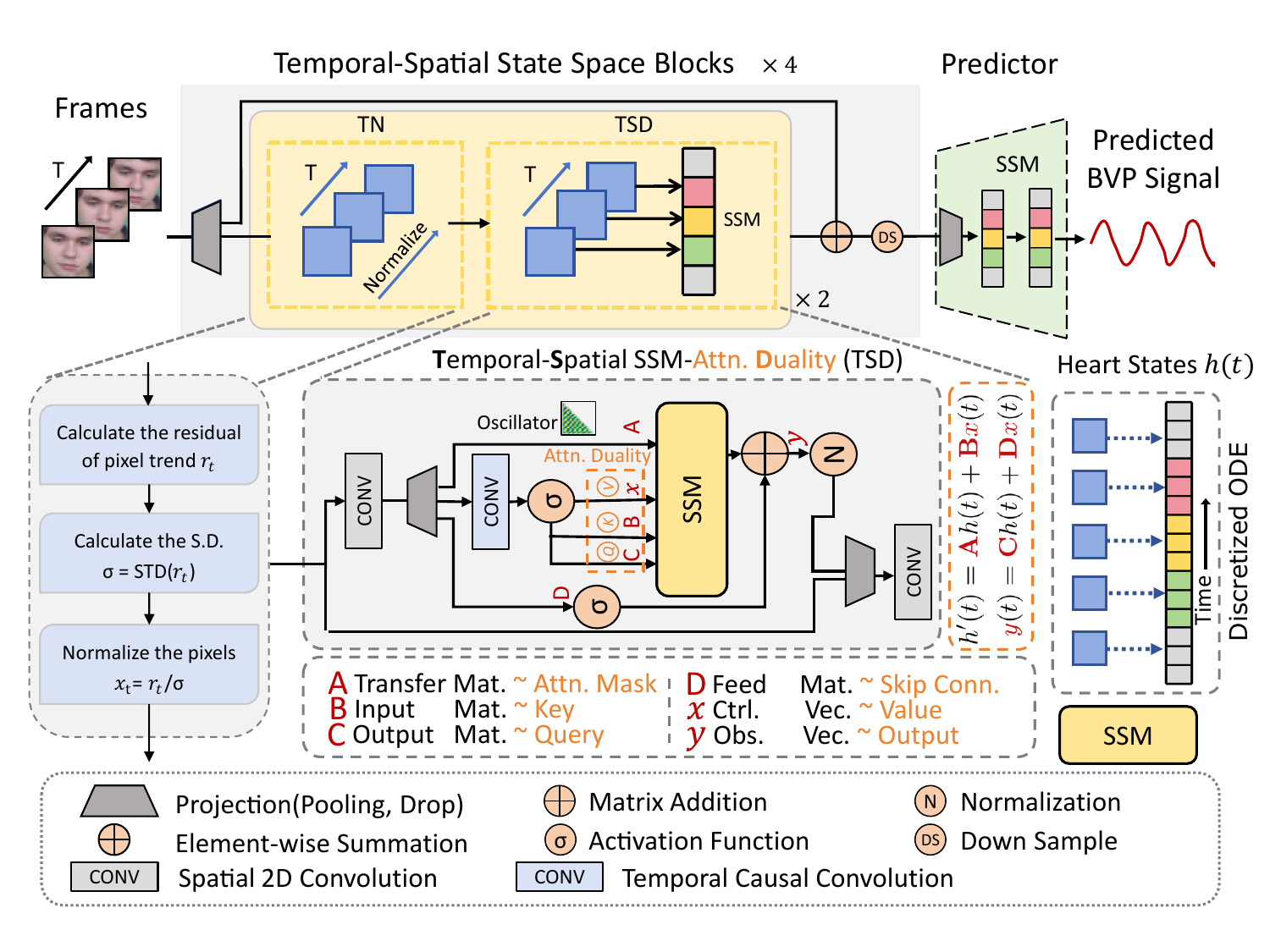}
    \vspace{-0.3cm}
    \caption{\textbf{FacePhys Framework.} Our framework utilizes the SSM dual as the core component, which serves both as an efficient discretization solver for the heart state LTI system and as a linear attention processor. It employs Temporal Normalization (TN) to stabilize the extraction of temporal features and introduces a complex state transition matrix A to enable periodic attention.}
    \label{fig:model}
    \vspace{-0.3em}
\end{figure*}
Our framework is designed as a discretized LTI solver, which is a state space model with a structured dual form.

Due to the three-dimensional nature of video, we employ temporal-spatial decoupled modeling rather than a simple patch embedding. During training, it inputs video frames and outputs the blood volume pulse (BVP) waveform $\mathbf{y} \in \mathbb{R}^{B \times T}$. During inference, the model supports streaming per-frame waveform prediction through recurrent state updates, where each frame produces one waveform sample. Heart rate is then computed from the predicted waveform using standard spectral analysis (FFT/PSD) over a temporal window.

The model processes video sequence tensors $\mathbf{X} \in \mathbb{R}^{B \times T \times H \times W \times C}$ as input and extracts BVP signals through temporal-spatial state space blocks. As illustrated in Figure~\ref{fig:model}, the computational workflow comprises two critical stages:
\begin{enumerate}
    \item Temporal Normalization (TN): Independently performs detrending and standardization across the temporal dimension for each spatial coordinate.
    \item Temporal-Spatial Modeling: Achieves periodic long-range temporal modeling through Temporal-Spatial SSM-Attention Duality (TSD).
\end{enumerate}
Unlike traditional spatio-temporal joint modeling, TSD decouples spatial and temporal dimensions, utilizing  2D convolutions for spatial modeling while employing SSM duality for temporal modeling. This separation enables memory-efficient inference, achieved efficient attention with linear computational complexity.
For a detailed comparison of memory efficiency with conventional architectures, refer to Table \ref{tab:memory complexity}.

\subsection{Temporal Normalization Module}

The TN module~\citep{wang2024plugandplaytemporalnormalizationmodule} ensures the stability of temporal characteristics in long sequences. It begins by calculating the linear trend term using the least squares method, as formulated in Equation \eqref{eq:detrend}. During training, the trend estimation can be computed in parallel.

\begin{equation}
(\beta_0, \beta_1) = \arg\min{\beta} \sum_{t=1}^T \left( x_t - (\beta_0 t + \beta_1) \right)^2 \label{eq:detrend}
\end{equation}

After obtaining the trend coefficients $(\beta_0, \beta_1)$, the residual of each pixel relative to the trend is computed and subsequently standardized as the final output.

$$
\begin{cases}
\text{Residual} & r_t = x_t - (\beta_0 t + \beta_1) \\
\text{S.D.} & \sigma = \sqrt{\frac{1}{T}\sum_{t=1}^T r_t^2} \\
\text{Output} & \tilde{x}_t = r_t / \sigma
\end{cases}
$$

To achieve constant time and space complexity during inference, the linear trend needs to be replaced with a Recursive Moving Average (RMA) trend.

$$
\begin{cases}
\text{RMA Trend} & \mu_t = \alpha \mu_{t-1} + (1-\alpha)x_t \\

\text{Residual} & r_t = x_t - \mu_t \\

\text{S.D.} & \sigma_t = \sqrt{\alpha \sigma_{t-1}^2 + (1-\alpha)r_t^2} \\
\text{Output} & \tilde{x}_t = r_t / \sigma_t
\end{cases}
$$

By updating the trend using RMA, RMA maintains fixed-size running statistics without buffering
the full input history. Moreover, RMA exhibits a strong similarity to the linear trend, making it an effective alternative for real-time processing.

\subsection{Complex Diagonal State Transition Matrix A}
\label{sec:cstm}

\begin{equation}
\begin{aligned}
h'(t) &= \mathbf{A}h(t) + \mathbf{B}x(t) \\
\end{aligned}
\label{eq:ODEtransfer}
\end{equation}

The oscillatory term introduces periodic representation, see Figure \ref{fig:matrix}. The solution to the controlled LTI system \eqref{eq:ODEtransfer} consists of a homogeneous solution and a particular solution, where the particular solution depends on the control term, while the homogeneous solution describes the dissipation and oscillation components during state transition, which are determined by the eigenvalues of the trainable matrix A. Since the control term $x(t)$ is governed by the input, we focus on the behavior of the homogeneous term. The cardiac pattern exhibits physiological periodicity, so matrix A should include oscillatory terms. However, since S4D~\citep{s4dgu2022parameterizationinitializationdiagonalstate}, many SSMs fix $A$ to a real diagonal form for efficiency; this removes complex eigenvalues and suppresses oscillations (Eq.~\eqref{eq:solution}). To match physiology without sacrificing speed, we keep the trainable diagonal structure but allow complex diagonal entries, which restores oscillatory behavior.

\vspace{-0.2em}
\begin{equation}
\begin{aligned}
h(t) = \sum_{j=1}^{k} c_j e^{\lambda_j t} v_j + \sum_{l=1}^{m} e^{\alpha_l t} \left[ A_l \cos(\beta_l t) + B_l \sin(\beta_l t) \right]
\end{aligned}
\label{eq:solution}
\end{equation}
\vspace{-0.2em} 

Where:
\( c_j \) are real constants (corresponding to real eigenvalues),\( A_l \) and \( B_l \) are real constant vectors (corresponding to pairs of complex eigenvalues), \( \alpha_l \) and \( \beta_l \) are the real and imaginary parts of the complex eigenvalues. Clearly, a real diagonal matrix A has no complex eigenvalues and cannot produce periodic behavior. However, if we remove the diagonal constraint on A, the computational complexity increases from linear to $O(n^3)$, which cannot meet efficiency requirements. Therefore, introducing complex diagonal elements is considered to improve this. As shown in Eq. \eqref{eq:global state}, the matrix L is equivalent to a causal mask, which describes the causality and decay behavior of state transitions. Its elements are determined by $L_{i,j} = \bar{A}^{i-j}$ when $i \geq j$, and the complex eigenvalues of A lead to periodic \( L_{i,j} \), thus it is dual to a periodic attention mask.
\vspace{-0.5em}
\section{Experiments}
\label{sec:experiment}
\vspace{-0.3em}

\subsection{Datasets}
\label{sec:dataset}
\vspace{-0.2em}
We use five benchmark datasets to evaluate our method: UBFC-rPPG~\citep{ubfcrppg}, PURE~\citep{pure}, MMPD~\citep{tang2023mmpd}, RLAP-rPPG~\citep{wang2023physbench}, and VitalVideo~\citep{Toye2023VitalVA}. UBFC-rPPG includes 43 RGB videos under controlled indoor lighting. PURE contains 59 videos with various head movements for motion robustness testing. MMPD comprises 660 videos under diverse lighting and activities. RLAP features 58 subjects with high synchronization. VitalVideo, the largest dataset, covers 893 subjects with varied skin tones and indoor lighting. These datasets were selected for accessibility, reliability, diversity, and synchronization quality, making them suitable for remote physiological signal extraction evaluation. All baselines are strictly reproduced according to the original papers and use the same training data and processing methods. FacePhys training details are provided in the supplementary material.

\begin{table*}[h]
    \vspace{-0.5em}
    \centering
    \caption{\textbf{Intra-Dataset Results.} FacePhys performed best across all datasets, when performing five-fold cross-validation.}
    \vspace{-0.3cm}
    \label{tab:intra}
    \begin{tabular}{lcccccccccc}
    \toprule[1.5pt]
    \multirow{2}{*}{Model} & \multicolumn{3}{c}{MMPD} & \multicolumn{3}{c}{VitalVideo} & \multicolumn{3}{c}{UBFC} \\
    \cmidrule(lr){2-4} \cmidrule(lr){5-7} \cmidrule(lr){8-10}
     & MAE & RMSE & R & MAE & RMSE & R & MAE & RMSE & R \\
    \midrule \midrule
    EFFPhys & 14.8 & 20.3 & 0.31 & 2.35 & 7.17 & 0.83 & 3.41 & 6.55 & 0.89 \\
    TSCAN & 14.2 & 19.4 & 0.39 & 2.93 & 7.54 & 0.83 & 2.36 & 4.08 & 0.98 \\
    PhysNet & 8.13 & 12.4 & 0.56 & 0.69 & \textbf{2.66} & \textbf{0.98} & 1.76 & 2.92 & 0.98 \\
    PhysFormer & 9.47 & 14.0 & 0.47 & 0.66 & 2.76 & \textbf{0.98} & 5.42 & 8.99 & 0.85 \\
    RhyMamba & 8.21 & 12.0 & 0.61 & 0.83 & 3.69 & 0.96 & 0.53 & 0.73 & \textbf{1.00} \\
    
    \rowcolor{gray!30}
    FacePhys & \textbf{5.58} & \textbf{10.2} & \textbf{0.72} & \textbf{0.64} & \textbf{2.66} & \textbf{0.98} & \textbf{0.46} & \textbf{0.67} & \textbf{1.00} \\ \hline
    \texttt{Gains (\%)} & \textcolor{darkgreen}{\texttt{+}\textbf{31.4}} & \textcolor{darkgreen}{\texttt{+}\textbf{15.0}} & \textcolor{darkgreen}{\texttt{+}\textbf{18.0}} & \textcolor{darkgreen}{\texttt{+}\textbf{3.0}} & \textcolor{darkgreen}{\texttt{+}\textbf{0.0}} & \textcolor{darkgreen}{\texttt{+}\textbf{0.0}} & \textcolor{darkgreen}{\texttt{+}\textbf{13.2}} & \textcolor{darkgreen}{\texttt{+}\textbf{8.2}} & \textcolor{darkgreen}{\texttt{+}\textbf{0.0}} \\
    
    \bottomrule[1.5pt]
    \end{tabular}
    \scriptsize \noindent \\MAE $\downarrow$ = Mean Absolute Error in HR estimation (Beats/Min). RMSE $\downarrow$ = Root Mean Squared Error in HR estimation (Beats/Min). R $\uparrow$ = Pearson Correlation in HR estimation.
    \end{table*}

\vspace{-0.3em}
\subsection{Comparison with State-of-the-Art Methods}
\label{sec:comparison}
\vspace{-0.2em}
\textbf{Intra-dataset Experiments.} We first compare our method with five SOTA methods (Efficient-Phys~\citep{efficientphys}, TSCAN~\citep{mttscan}, PhysNet~\citep{physnet}, PhysFormer~\citep{physformer}, RhythmMamba~\citep{zou2024rhythmmamba}) on three representative datasets (MMPD~\citep{tang2023mmpd}, VitalVideo~\citep{Toye2023VitalVA}, and UBFC~\citep{ubfcrppg}) using 5-fold subject-independent intra-dataset settings. The results are shown in Table~\ref{tab:intra}. Our method outperforms all other methods on almost all datasets, demonstrating its effectiveness in remote PPG and HR estimation. Specifically, FacePhys achieves the lowest MAE of 5.58 on MMPD and the highest Pearson Correlation of 1.00 on UBFC dataset, highlighting its superior accuracy and reliability.

\begin{table*}[h!]
    \vspace{-0.5em}
    \centering
    \caption{\textbf{Cross-dataset results.} FacePhys performs best in all comparisons (trained on \textbf{RLAP}).}
    \vspace{-0.3cm}
    \label{tab:cross_rlap}
    \begin{tabular}{lccccccccc}
    \toprule[1.5pt]
    \multirow{2}{*}{Model} & \multicolumn{3}{c}{MMPD} & \multicolumn{3}{c}{VitalVideo} & \multicolumn{3}{c}{PURE} \\
     \cmidrule(lr){2-4} \cmidrule(lr){5-7} \cmidrule(lr){8-10}
     & MAE & RMSE & R & MAE & RMSE & R & MAE & RMSE & R \\
    \midrule \midrule
    EFFPhys  & 9.18 & 13.7 & 0.56 & 6.75 & 19.3 & 0.51 & 2.48 & 5.48 & 0.98 \\
    TSCAN & 10.2 & 15.2 & 0.47 & 4.20 & 8.86 & 0.78 & 3.90 & 7.05 & 0.96 \\
    PhysNet  & 11.2 & 16.4 & 0.42 & 0.89 & 3.56 & 0.96 & 0.63 & 2.34 & 1.00 \\
    PhysFormer  & 12.8 & 18.9 & 0.28 & 4.97 & 10.1 & 0.70 & 0.91 & 3.32 & 0.99 \\
    RhyMamba & 9.62 & 14.9 & 0.47 & 2.85 & 8.91 & 0.80 & 1.63 & 4.55 & 0.98 \\
    
    \rowcolor{gray!30}
    FacePhys  & \textbf{5.30} & \textbf{10.0} & \textbf{0.76} & \textbf{0.77} & \textbf{3.07} & \textbf{0.97} & \textbf{0.24} & \textbf{0.68} & \textbf{1.00} \\ \hline
    \texttt{Gains} & \textcolor{darkgreen}{\texttt{+}\textbf{42.3}} & \textcolor{darkgreen}{\texttt{+}\textbf{27.0}} & \textcolor{darkgreen}{\texttt{+}\textbf{35.7}} & \textcolor{darkgreen}{\texttt{+}\textbf{13.5}} & \textcolor{darkgreen}{\texttt{+}\textbf{13.8}} & \textcolor{darkgreen}{\texttt{+}\textbf{1.0}} & \textcolor{darkgreen}{\texttt{+}\textbf{61.9}} & \textcolor{darkgreen}{\texttt{+}\textbf{70.9}} & \textcolor{darkgreen}{\texttt{+}\textbf{2.0}} \\
    
    \bottomrule[1.5pt]
    \end{tabular}
    \scriptsize \noindent \\MAE $\downarrow$ = Mean Absolute Error in HR estimation (Beats/Min). RMSE $\downarrow$ = Root Mean Squared Error in HR estimation (Beats/Min). R $\uparrow$ = Pearson Correlation in HR estimation.
    \end{table*}
    
    \begin{table*}[h!]
    \vspace{-0.5em}
    \centering
    \caption{\textbf{Cross-dataset results.} FacePhys performs best in all comparisons (trained on \textbf{PURE}).}
    \vspace{-0.3cm}
    \label{tab:cross_pure}
    \begin{tabular}{lccccccccc}
    \toprule[1.5pt]
    \multirow{2}{*}{Model} & \multicolumn{3}{c}{MMPD} & \multicolumn{3}{c}{VitalVideo} & \multicolumn{3}{c}{UBFC} \\
     \cmidrule(lr){2-4} \cmidrule(lr){5-7} \cmidrule(lr){8-10}
    & MAE & RMSE & R & MAE & RMSE & R & MAE & RMSE & R \\
    \midrule \midrule
    EFFPhys  & 17.6 & 23.0 & 0.26 & 3.83 & 8.42 & 0.79 & 2.96 & 5.11 & 0.97 \\
    TSCAN  & 18.1 & 23.8 & 0.22 & 3.24 & 8.03 & 0.82 & 2.71 & 5.70 & 0.95 \\
    PhysNet  & 15.1 & 20.5 & 0.29 & 1.32 & 5.13 & 0.93 & 1.62 & 3.49 & 0.98 \\
    PhysFormer  & 13.0 & 18.1 & 0.44 & 1.08 & 4.07 & 0.95 & 1.15 & 2.06 & 0.99 \\
    RhyMamba  & 10.9 & 15.2 & 0.42 & 1.74 & 5.87 & 0.91 & 0.63 & 1.07 & \textbf{1.00} \\
    
    \rowcolor{gray!30}
    FacePhys  & \textbf{8.45} & \textbf{13.1} & \textbf{0.61} & \textbf{0.90} & \textbf{3.37} & \textbf{0.97} & \textbf{0.48} & \textbf{0.72} & \textbf{1.00} \\  \hline
    \texttt{Gains} & \textcolor{darkgreen}{\texttt{+}\textbf{22.5}} & \textcolor{darkgreen}{\texttt{+}\textbf{13.8}} & \textcolor{darkgreen}{\texttt{+}\textbf{38.6}} & \textcolor{darkgreen}{\texttt{+}\textbf{16.7}} & \textcolor{darkgreen}{\texttt{+}\textbf{17.2}} & \textcolor{darkgreen}{\texttt{+}\textbf{2.1}} & \textcolor{darkgreen}{\texttt{+}\textbf{23.8}} & \textcolor{darkgreen}{\texttt{+}\textbf{32.7}} & \textcolor{darkgreen}{\texttt{+}\textbf{0.0}} \\
    
    \bottomrule[1.5pt]
    \end{tabular}
    \scriptsize \noindent \\MAE $\downarrow$ = Mean Absolute Error in HR estimation (Beats/Min). RMSE $\downarrow$ = Root Mean Squared Error in HR estimation (Beats/Min). R $\uparrow$ = Pearson Correlation in HR estimation.
    \end{table*}

\textbf{Cross-dataset Experiments.} To assess the generalization ability of our method, we evaluate its performance in cross-dataset settings. The results in Tables~\ref{tab:cross_rlap} and ~\ref{tab:cross_pure} indicate that FacePhys achieves the best performance across \emph{all} the cross-dataset settings, demonstrating strong generalization. Specifically, FacePhys achieves the lowest MAE of 5.30 and the highest R of 0.76 on the challenging MMPD dataset when trained on the RLAP dataset.  FacePhys outperforms all methods on the largest VitalVideo dataset (MAE: 0.77, R: 0.97) when trained on RLAP, confirming its robustness and reliability. Compared to cross-dataset results, FacePhys maintains consistent performance across different datasets, whereas other methods show significant performance drops and less consistent performance. This demonstrates our method's effectiveness in handling diverse real-world scenarios and generalizing well across datasets.

\textbf{Computational Efficiency.} We further assess the computational efficiency of our method by comparing its inference time against other SOTA approaches. As shown in Table~\ref{tab:efficiency}, FacePhys achieves the best trade-off between accuracy and efficiency, underscoring its practicality for real-world applications. FacePhys achieves the lowest latency of 9.46 ms, which is even lower than the frame capture time of 33.3 ms at 30 fps, making it suitable for real-time PPG estimation on resource-constrained devices.

\begin{wraptable}{r}{0.6\textwidth}
\vspace{-2em}
\centering
\caption{\textbf{Training and Test Chunk Length Ablation.} \\ Longer video chunks lead to lower HR estimation error.}
\vspace{-0.3cm}
\label{tab:chunk}
\begin{tabular}{cccccc}
\toprule[1.5pt]
Train\textbackslash Test & 160 & 320 & 640 & 1280 & 1800 \\
\midrule \midrule
160 & 6.13 & 6.05 & 6.11 & 6.15 & 6.11 \\
320 & 6.17 & 6.19 & 6.17 & 6.09 & 5.78 \\
640 & 6.24 & 6.05 & 6.02 & 5.86 & 5.79 \\
1280 & 5.92 & 5.91 & 5.80 & 5.66 & 5.75 \\
1600 & 5.93 & 5.73 & 5.51 & 5.42 & \textbf{5.30} \\
\bottomrule[1.5pt]
\end{tabular}
\scriptsize \noindent \\ Experiments are trained on RLAP and tested on MMPD. MAE is calculated in HR estimation (Beats/Min). Each video is divided into chunks of different lengths.
\vspace{-1.5em}
\end{wraptable}

\vspace{-0.3em}
\subsection{Ablation Studies}
\label{sec:ablation}
\vspace{-0.2em}


\textbf{Ablation Study on Components.} We conduct an ablation study to validate the effectiveness of each component in FacePhys. We compare the performance after removing different components, including TN, SSD modules and oscillator matrix A. The results in Table~\ref{tab:ablation} show that our method performs best when all components are included. The TN module maintains the stability of temporal features, while the SSD module provides attention to assign weights to different frames. After the trainable state transition matrix A is replaced with complex values, oscillatory terms emerge, which enhances the model's ability to represent periodicity.

\begin{table*}[ht]
\vspace{-0.5em}
\centering
\caption{\textbf{Architecture Ablation.} Removing the Temporal Normalization, State Space or Oscillator Matrix A will harm performance.}
\vspace{-0.3cm}
\label{tab:ablation}
\begin{tabular}{rccccccccc}
\toprule[1.5pt]
\multirow{2}{*}{Model} & \multicolumn{3}{c}{MMPD} & \multicolumn{3}{c}{VitalVideo} & \multicolumn{3}{c}{PURE} \\
\cmidrule(lr){2-4} \cmidrule(lr){5-7} \cmidrule(lr){8-10}
 & MAE & RMSE & R & MAE & RMSE & R & MAE & RMSE & R \\
\midrule \midrule
w/o TN & 11.6 & 17.6 & 0.43 & 0.88 & 3.45 & 0.97 & 0.80 & 3.97 & 0.99 \\
w/o SSM Duality & 10.8 & 16.7 & 0.51 & 1.10 & 3.80 & 0.96 & 0.34 & 0.87 & 1.00 \\
w/o Oscillator & 7.04 & 11.7 & 0.71 & 0.80 & 3.08 & 0.97 & 0.26 & 0.72 & 1.00 \\
\rowcolor{gray!30}
FacePhys & \textbf{5.30} & \textbf{10.0} & \textbf{0.76} & \textbf{0.77} & \textbf{3.07} & \textbf{0.97} & \textbf{0.24} & \textbf{0.68} & \textbf{1.00} \\ \hline
\texttt{Gains} & \textcolor{darkgreen}{\texttt{+}\textbf{24.5}} & \textcolor{darkgreen}{\texttt{+}\textbf{14.5}} & \textcolor{darkgreen}{\texttt{+}\textbf{7.0}} & \textcolor{darkgreen}{\texttt{+}\textbf{3.7}} & \textcolor{darkgreen}{\texttt{+}\textbf{0.3}} & \textcolor{darkgreen}{\texttt{+}\textbf{0.0}} & \textcolor{darkgreen}{\texttt{+}\textbf{7.7}} & \textcolor{darkgreen}{\texttt{+}\textbf{5.6}} & \textcolor{darkgreen}{\texttt{+}\textbf{0.0}} \\
\bottomrule[1.5pt]
\end{tabular}
\scriptsize \noindent \\MAE $\downarrow$ = Mean Absolute Error in HR estimation (Beats/Min). RMSE $\downarrow$ = Root Mean Squared Error in HR estimation (Beats/Min). R $\uparrow$ = Pearson Correlation in HR estimation.
\end{table*}

\textbf{Ablation Study on Training and Test Chunk Length.} We also examine the impact of different training and test chunk lengths on model performance. The results in Table~\ref{tab:chunk} show that longer chunks lead to better performance by capturing more temporal information. This finding aligns with previous studies~\citep{zou2024rhythmmamba} that longer test lengths can improve rPPG performance. With our memory-efficient design, we can train on entire videos, further proving that longer training lengths also enhance performance. Our method achieves the best results with a training chunk length of 1600 and a test chunk length of 1800, demonstrating its effectiveness in handling long-term temporal dependencies.

    \vspace{-0.2cm}
\subsection{Temporal Analysis}
\label{sec:temporal_analysis}
    \vspace{-0.2cm}
    
\begin{figure}[b]
    \centering
    \includegraphics[width=0.78\linewidth]{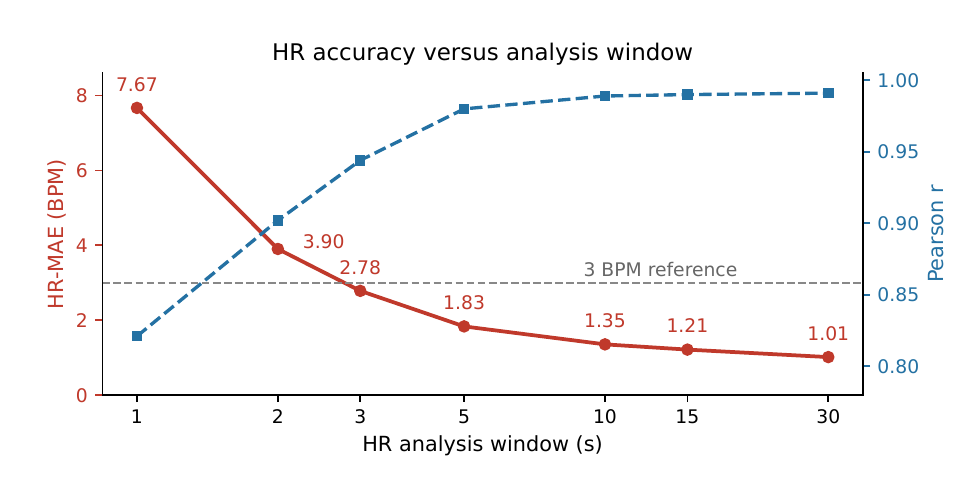}
    \vspace{-0.3cm}
    \caption{\textbf{Effect of HR analysis window length.}
    Longer windows improve HR estimation accuracy on UBFC,
    with diminishing gains.}
    \label{fig:hr_window}
\end{figure}

\textbf{Effect of HR estimation window length.}
We examine how the duration of the HR analysis window affects estimation
accuracy using saved BVP predictions from RLAP-trained FacePhys on UBFC.
Unlike the training and test chunk lengths in Table~\ref{tab:chunk},
this experiment varies the temporal window used for spectral HR estimation.
As shown in Figure~\ref{fig:hr_window}, HR MAE decreases
from 7.67 BPM at 1 s to 2.78 BPM at 3 s and 1.83 BPM at 5 s.
Longer windows further reduce the error to 1.21 BPM at 15 s and
1.01 BPM at 30 s, with diminishing improvements.
These results characterize the trade-off between observation duration
and HR accuracy. Although FacePhys produces BVP samples frame by frame,
HR estimation requires accumulating a temporal window.
The reported 9.46 ms per-frame inference time therefore describes
waveform computation rather than the time required to obtain a stable
HR estimate.

\begin{figure}[t]
    \centering
    \includegraphics[width=0.8\linewidth]{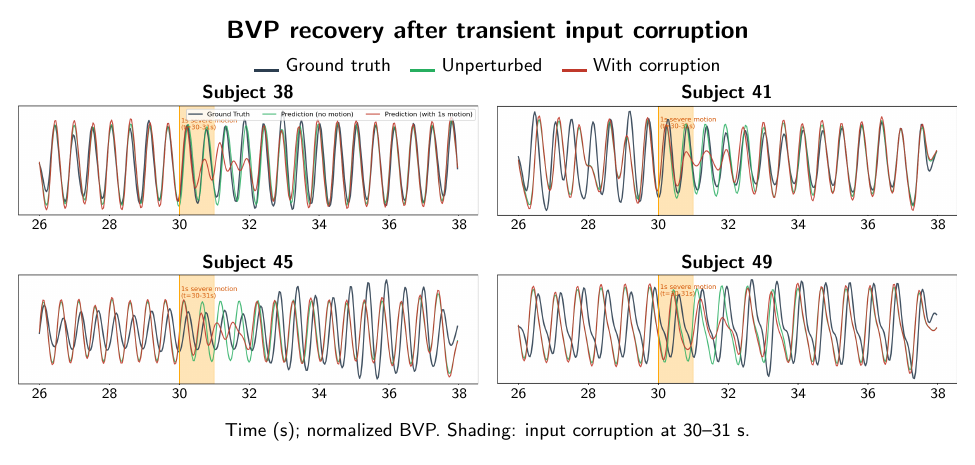}
    \caption{\textbf{Recovery after transient input corruption.}
    Across 12 UBFC subjects, recovery takes 0.92 s on average
    (maximum: 1.4 s), measured by correlation $\geq 0.9$
    with the unperturbed output.}
    \vspace{-0.3cm}
    \label{fig:recovery}
\end{figure}

\textbf{Recovery after transient input corruption.}
To assess whether a brief disturbance causes persistent errors during
streaming inference, we evaluate RLAP-trained FacePhys on 12 UBFC
subjects. We replace the input frames between 30 and 31 s with
random-noise images and then resume the original video.
Recovery time is measured from the end of the perturbation until the
predicted BVP regains a correlation of at least 0.9 with the output
of an unperturbed run.
The mean recovery time is 0.92 s, with a maximum of 1.4 s.
Figure~\ref{fig:recovery} illustrates the output trajectories
for four subjects. The perturbed predictions return toward their
unperturbed trajectories after clean input resumes, supporting recovery
from transient corruption in this controlled setting.
This experiment evaluates output recovery under synthetic corruption;
it does not directly measure hidden-state convergence or establish
robustness to arbitrary real-world motion.

\vspace{-0.5em}
\section{Discussion}
\label{sec:discussion}
\vspace{-0.3em}

\subsection{An Effective and Generalizable Model}
\vspace{-0.2em}
\begin{wrapfigure}{r}{0.5\textwidth}
    \vspace{-1.5em}
    \centering
    \includegraphics[width=\linewidth]{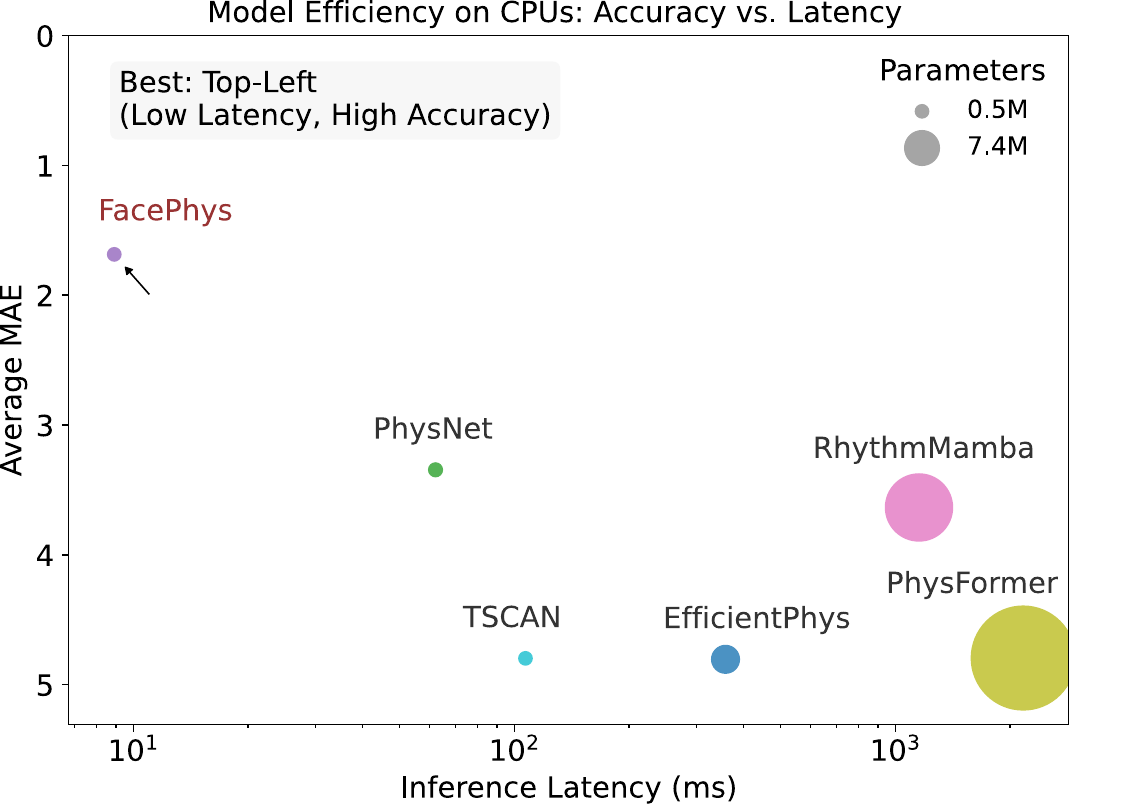}
    \vspace{-0.5em}
    \caption{In terms of model accuracy and latency, FacePhys's efficiency performs best.}
    \label{fig:bubble}
\end{wrapfigure}

FacePhys achieves state-of-the-art performance in both intra-dataset and cross-dataset testing, excelling on challenging MMPD and large-scale VitalVideo datasets compared to nearly saturated PURE and UBFC. In cross-dataset validation, FacePhys maintains consistent performance while other methods degrade significantly. When trained on RLAP, MAE on VitalVideo remains below 1 BPM (0.77), with MMPD achieving 5.30 BPM error. When trained on PURE, MAE on both VitalVideo (0.90) and UBFC (0.48) also remains below 1 BPM, demonstrating strong performance.

This robust generalization stems from FacePhys's TN and TSD design, which embeds physiological priors and decoupled spatial and temporal modeling, distinguishing it from prior Mamba works as a physiology-driven SSM. Studies~\citep{physnet} found longer videos did not benefit CNNs due to limited receptive fields, while Table~\ref{tab:memory complexity} shows Transformers incur impractical quadratic costs with increased input length. FacePhys efficiently captures long-term dependencies with minimal overhead, enabling training on extended segments and enhancing cross-dataset generalization.

\vspace{-0.3em}
\subsection{Efficient and Real-time FacePhys}
\vspace{-0.2em}

Real-time performance is crucial for medical monitoring and video streaming applications. Traditional methods like Green~\citep{verkruysse2008green} process single frames but suffer from motion artifacts, while deep learning methods improve accuracy but require buffering long segments, limiting real-time use. Models like RTrPPG~\citep{rtrppg} claim real-time inference yet require pre-recorded input, and TS-CAN~\citep{liu2020multi} requires 30-second recording before output. With 719K parameters, FacePhys uses 3.6 MB memory and achieves 9.46 ms per-frame latency, enabling streaming waveform inference on mobile devices.

FacePhys's efficiency stems from its streaming inference paradigm: unlike methods constructing temporal dependencies within fixed segments, FacePhys maintains a transferable heart state updated frame-by-frame through recurrent transitions, outputting one waveform sample per frame while accumulating forward information for stable predictions. Heart rate is computed from the waveform via spectral analysis (FFT/PSD) over a temporal window, introducing unavoidable lag inherent to frequency-based HR measurement. The lightweight TN and TSD design enable this streaming architecture for time-sensitive and resource-limited applications. The temporal analyses in Section~\ref{sec:temporal_analysis} further
show that streaming waveform inference can recover from brief
synthetic input corruption, while HR accuracy remains dependent on
the duration of the spectral analysis window.

\vspace{-0.3em}
\subsection{Limitations and Future Work}
\vspace{-0.2em}

Although FacePhys has made significant progress in performance and real-time capabilities, certain limitations remain. First, to ensure single frame inference, our model cannot access future information; this introduces lag. Second, although we validated our model on large scale diverse datasets, we have not yet tested it in hospital clinical scenarios, where performance for users with cardiovascular diseases remains to be evaluated.

Future work can extend FacePhys to predict additional cardiovascular signals from facial videos, including blood oxygen~\citep{tang2025camera} and blood pressure~\citep{curran2023camera}. Enhancing rPPG accuracy through multimodal fusion with thermal imaging or IMU sensors is another potential direction~\citep{ma2025non}. Optimizing computational efficiency will enable real-time processing at higher frame rates on lightweight hardware, expanding its practical applications in remote healthcare monitoring.

\vspace{-0.3em}
\subsection{Broader Impacts}
\vspace{-0.2em}

FacePhys enables accessible cardiovascular monitoring through low-cost deployment on consumer devices, democratizing healthcare for underserved populations. On-device processing minimizes data transmission, enhancing privacy protection compared to cloud-based solutions. However, facial video capture raises privacy concerns, requiring explicit user consent and transparent data policies. The technology could be misused for unauthorized surveillance, necessitating ethical frameworks and regulatory compliance. As rPPG cannot replace clinical-grade devices, systems must clearly communicate uncertainty and recommend physician consultation for diagnosis. Performance may vary across skin tones and demographics; we advocate rigorous subgroup validation and continuous bias monitoring to ensure equitable healthcare access across diverse populations.

\vspace{-0.5em}
\section{Conclusion}
\label{sec:conclusion}
\vspace{-0.3em}

In this paper, we propose FacePhys, a memory-efficient rPPG algorithm based on temporal-spatial state space duality, designed for arbitrary length video training and inference. Our approach effectively captures subtle temporal changes and spatial representations of facial vascular information, achieving significant improvements over state-of-the-art methods on large-scale datasets such as MMPD and VitalVideo. FacePhys demonstrates superior accuracy and memory efficiency, with improvements of up to 60.3\% in accuracy and 98.4\% in memory efficiency. To the best of our knowledge, FacePhys is the first rPPG model that supports streaming waveform inference through recurrent
state updates, reducing computational overhead. Our method provides new directions for future research on rPPG, effectively balancing accuracy and efficiency for practical deployment.

\vspace{-0.5em}
\section*{Acknowledge}
\label{sec:acknowledge}
\vspace{-0.3em}

This work is supported by the Space Medical Experiment Project of CMSP under grant No. HYZHXMR01017,  the National Natural Science Foundation of China under Grant No. 62472244, the foundation of National Key Laboratory of Human Factors Engineering under Grant No. HFNKL2024W06, the Tsinghua University Initiative Scientific Research Program under Grant No.20257020004, and Qinghai University Research Ability Enhancement Project under Grant No. 2025KTSA05.

\bibliographystyle{ieeenat_fullname}
\bibliography{main}


\appendix



\newpage
\section*{NeurIPS Paper Checklist}

\begin{enumerate}

\item {\bf Claims}
    \item[] Question: Do the main claims made in the abstract and introduction accurately reflect the paper's contributions and scope?
    \item[] Answer: \answerYes{}
    \item[] Justification: The abstract and introduction accurately describe FacePhys as a memory-efficient rPPG algorithm based on LTI state space models with SSD, achieving 49\% error reduction and real-time streaming inference (3.6 MB memory, 9.46 ms latency), which are validated in the experiments section.

\item {\bf Limitations}
    \item[] Question: Does the paper discuss the limitations of the work performed by the authors?
    \item[] Answer: \answerYes{}
    \item[] Justification: Section 5.3 (Limitations and Future Work) discusses limitations including inability to access future information for single frame inference, lack of clinical validation in hospital scenarios, and untested performance for users with cardiovascular diseases.

\item {\bf Theory assumptions and proofs}
    \item[] Question: For each theoretical result, does the paper provide the full set of assumptions and a complete (and correct) proof?
    \item[] Answer: \answerNA{}
    \item[] Justification: This paper is primarily empirical, presenting a novel architecture for rPPG estimation. While it describes the mathematical formulation of the state space model (Eq. 2-6), it does not claim formal theoretical proofs or theorems.

    \item {\bf Experimental result reproducibility}
    \item[] Question: Does the paper fully disclose all the information needed to reproduce the main experimental results of the paper to the extent that it affects the main claims and/or conclusions of the paper (regardless of whether the code and data are provided or not)?
    \item[] Answer: \answerYes{}
    \item[] Justification: The paper describes the architecture in detail (Section 3), datasets (Section 4.1), and states that training details are in supplementary material. The abstract mentions training code is provided in supplementary materials.

\item {\bf Open access to data and code}
    \item[] Question: Does the paper provide open access to the data and code, with sufficient instructions to faithfully reproduce the main experimental results, as described in supplemental material?
    \item[] Answer: \answerYes{}
    \item[] Justification: The abstract states "The training code and real-time inference demo for mobile browsers can be found in the supplementary materials." All datasets used are publicly available benchmark datasets.

\item {\bf Experimental setting/details}
    \item[] Question: Does the paper specify all the training and test details (e.g., data splits, hyperparameters, how they were chosen, type of optimizer) necessary to understand the results?
    \item[] Answer: \answerYes{}
    \item[] Justification: Section 4.1 describes datasets and evaluation protocol. The paper states that training details including data splits, hyperparameters, and experimental configurations are provided in the supplementary material.

\item {\bf Experiment statistical significance}
    \item[] Question: Does the paper report error bars suitably and correctly defined or other appropriate information about the statistical significance of the experiments?
    \item[] Answer: \answerNo{}
    \item[] Justification: The paper reports mean performance metrics (MAE, RMSE, Pearson R) across 5-fold cross-validation but does not include error bars or standard deviations. Statistical significance testing would be computationally expensive for the large-scale experiments conducted.

\item {\bf Experiments compute resources}
    \item[] Question: For each experiment, does the paper provide sufficient information on the computer resources (type of compute workers, memory, time of execution) needed to reproduce the experiments?
    \item[] Answer: \answerYes{}
    \item[] Justification: The supplementary material provides detailed compute resource information including hardware specifications, software environment, and training configuration. Table 1 in the paper reports inference latency and memory usage.
    
\item {\bf Code of ethics}
    \item[] Question: Does the research conducted in the paper conform, in every respect, with the NeurIPS Code of Ethics \url{https://neurips.cc/public/EthicsGuidelines}?
    \item[] Answer: \answerYes{}
    \item[] Justification: The research uses publicly available benchmark datasets for rPPG estimation and discusses ethical considerations including privacy, consent, and fairness in Section 5.4 (Broader Impacts).

\item {\bf Broader impacts}
    \item[] Question: Does the paper discuss both potential positive societal impacts and negative societal impacts of the work performed?
    \item[] Answer: \answerYes{}
    \item[] Justification: Section 5.4 (Broader Impacts) discusses positive impacts (democratizing healthcare, privacy-preserving on-device processing) and negative impacts (privacy concerns from facial video capture, potential surveillance misuse, performance variations across demographics, limitations vs clinical-grade devices).
    
\item {\bf Safeguards}
    \item[] Question: Does the paper describe safeguards that have been put in place for responsible release of data or models that have a high risk for misuse (e.g., pre-trained language models, image generators, or scraped datasets)?
    \item[] Answer: \answerNA{}
    \item[] Justification: The rPPG model for heart rate estimation from facial videos does not pose high misuse risks comparable to generative models. Section 5.4 discusses ethical considerations and recommends user consent and transparent data policies.

\item {\bf Licenses for existing assets}
    \item[] Question: Are the creators or original owners of assets (e.g., code, data, models), used in the paper, properly credited and are the license and terms of use explicitly mentioned and properly respected?
    \item[] Answer: \answerYes{}
    \item[] Justification: All datasets (UBFC-rPPG, PURE, MMPD, RLAP-rPPG, VitalVideo) are properly cited in Section 4.1 and were obtained through author permissions on request. Baseline methods are properly cited throughout the paper.

\item {\bf New assets}
    \item[] Question: Are new assets introduced in the paper well documented and is the documentation provided alongside the assets?
    \item[] Answer: \answerYes{}
    \item[] Justification: The paper introduces the FacePhys model with detailed architecture documentation in Section 3. Training code and real-time inference demo are provided in supplementary materials as stated in the abstract.

\item {\bf Crowdsourcing and research with human subjects}
    \item[] Question: For crowdsourcing experiments and research with human subjects, does the paper include the full text of instructions given to participants and screenshots, if applicable, as well as details about compensation (if any)?
    \item[] Answer: \answerNA{}
    \item[] Justification: The paper evaluates exclusively on publicly available benchmark datasets and does not involve crowdsourcing or new human subjects research.

\item {\bf Institutional review board (IRB) approvals or equivalent for research with human subjects}
    \item[] Question: Does the paper describe potential risks incurred by study participants, whether such risks were disclosed to the subjects, and whether Institutional Review Board (IRB) approvals (or an equivalent approval/review based on the requirements of your country or institution) were obtained?
    \item[] Answer: \answerNA{}
    \item[] Justification: The paper only uses publicly available benchmark datasets and does not involve new human subjects research requiring IRB approval.

\item {\bf Declaration of LLM usage}
    \item[] Question: Does the paper describe the usage of LLMs if it is an important, original, or non-standard component of the core methods in this research? Note that if the LLM is used only for writing, editing, or formatting purposes and does \emph{not} impact the core methodology, scientific rigor, or originality of the research, declaration is not required.
    \item[] Answer: \answerNA{}
    \item[] Justification: The core methodology of FacePhys is based on state space models and does not involve LLMs as a component. Any LLM usage would be limited to writing assistance, which does not require declaration.

\end{enumerate}

\end{document}